\documentclass[conference]{IEEEtran}
\IEEEoverridecommandlockouts
\usepackage{cite}
\usepackage{amsmath,amssymb,amsfonts}
\usepackage{algorithmic}
\usepackage{graphicx}
\usepackage{textcomp}
\usepackage{xcolor}
\usepackage{multirow}
\def\BibTeX{{\rm B\kern-.05em{\sc i\kern-.025em b}\kern-.08em
    T\kern-.1667em\lower.7ex\hbox{E}\kern-.125emX}}
\begin{document}

\title{An Exceptional Dataset For Rare Pancreatic Tumor Segmentation
}

\author{
    \IEEEauthorblockN{Wenqi Li$^{1\ast}$
    $\qquad$ Yingli Chen$^{2\ast}$ 
    $\qquad$ Keyang Zhou$^{1}$
    $\qquad$ Xiaoxiao Hu$^{1}$
    $\qquad$ Zilu Zheng$^{2}$
    $\qquad$ Yue	Yan$^{3}$\\
    $\qquad$ Xinpeng	Zhang$^{1}$
    $\qquad$ Wei Tang$^{2\dagger}$
    $\qquad$ Zhenxing Qian$^{1\dagger}$\vspace{0.3cm}}
    \IEEEauthorblockA{
        $^{1}$ School of Computer Science, Fudan University, Shanghai China \vspace{0.1cm}\\
        % \vspace{0.15cm}
        $^{2}$Shanghai Medical College, Fudan University Shanghai Cancer Center, Shanghai, China \vspace{0.1cm}\\
        % \vspace{0.1cm}
        $^{3}$School of Computer Science, University of Illinois, Illinois, United States
    }
    \thanks{
    $^\ast$These authors contributed equally to this work. $^\dag$ Corresponding authors.
This work is supported by the National Natural Science Foundation of China (82471933), and the Clinical Research Special Project of Shanghai Municipal Health Commission (202340123).
    }
   
}

\IEEEpubid{\begin{minipage}{\textwidth}
\vspace{0.7cm} 
\centering 
Copyright \copyright 20XX IEEE Personal use of this material is permitted. Permission from IEEE must be obtained for all other uses, in any current or future media, including reprinting/republishing this material for advertising or promotional purposes, creating new collective works, for resale or redistribution to servers or lists, or reuse of any copyrighted component of this work in other works.
\end{minipage}}

\maketitle

% \renewcommand{\thefootnote}{\fnsymbol{footnote}} %将脚注符号设置为fnsymbol类型，即特殊符号表示
% \footnotetext[1]
% {  These authors contributed equally to this work. $^\dag$ Corresponding authors.
% This work is supported by the National Natural Science Foundation of China (82471933), and the Clinical Research Special Project of Shanghai Municipal Health Commission (202340123).} %对应脚注[1]
% % \footnotetext[2]{} %对应脚注[2]

\begin{abstract}
Pancreatic NEuroendocrine Tumors (pNETs) are very rare endocrine neoplasms that account for less than 5\% of all pancreatic malignancies, with an incidence of only 1–1.5 cases per 100,000. Early detection of pNETs is critical for improving patient survival, but the rarity of pNETs makes segmenting them from CT a very challenging problem. So far, there has not been a dataset specifically for pNETs available to researchers. To address this issue, we propose a pNETs dataset, a well-annotated Contrast-Enhanced Computed Tomography (CECT) dataset focused exclusively on Pancreatic Neuroendocrine Tumors, containing data from 469 patients. This is the first dataset solely dedicated to pNETs, distinguishing it from previous collections. Additionally, we provide the baseline detection networks with a new slice-wise weight loss function designed for the UNet-based model, improving the overall pNET segmentation performance.
We hope that our dataset can enhance the understanding and diagnosis of pNET Tumors within the medical community, facilitate the development of more accurate diagnostic tools, and ultimately improve patient outcomes and advance the field of oncology.

\end{abstract}


\begin{figure}[t]
\centering
\includegraphics[width=1.0\columnwidth]{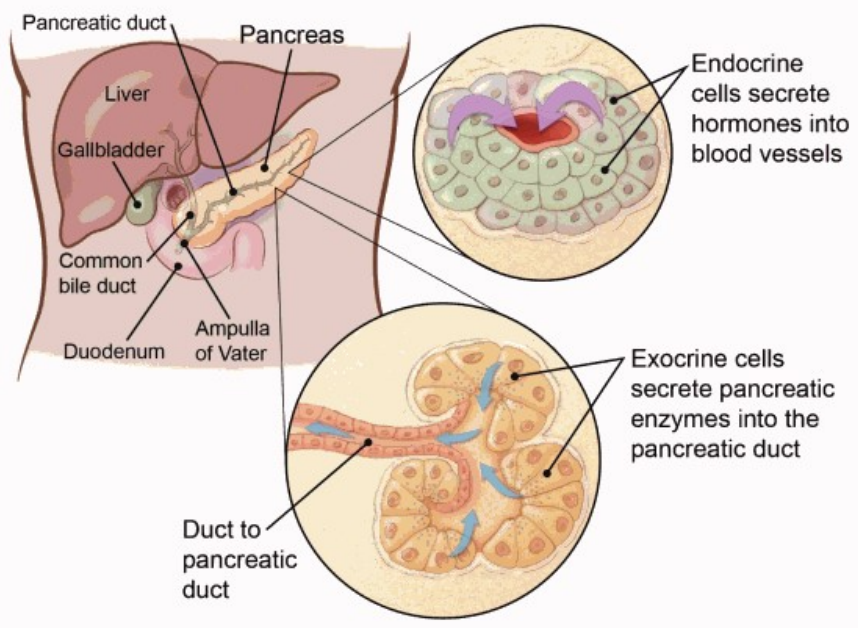} % Reduce the figure size so that it is slightly narrower than the column. Don't use precise values for figure width.This setup will avoid overfull boxes.
\caption{Demonstration of pancreas in human body\cite{ACS2024}. Pancreas consists both endocrine cells and exocrine cells. Most pancreatic cancers are characterised as ductal adenocarcinoma and thus represent malignancy of the exocrine pancreas whereas a minority represent neuroendocrine tumours\cite{mizrahi2020pancreatic}. Our work focus on this less populated type of pancreatic cancer.}
\label{pancreas_fig}
\end{figure}

\section{Introduction}

Pancreatic cancer has the poorest prognosis among solid tumor cancers, with a 5-year overall survival (OS) rate of around 10\%, positioning it as the second leading cause of cancer-related deaths in Western countries by 2030\cite{mizrahi2020pancreatic}. 
% This grim outlook is largely due to the lack of early, disease-specific symptoms, leading to the diagnosis of 80–85\% of patients at advanced stages\cite{siegel2024cancer}.
Surgical resection remains the only potentially curative treatment, although a low percentage of patients are eligible due to the advanced stage of diagnosis, highlighting the critical importance of accurate preoperative assessment of surgical resectability for improving patient survival\cite{wood2022pancreatic}.
% As Figure \ref{pancreas_fig} shows, the pancreas is an organ that sits behind the stomach. It's shaped a bit like a fish with a wide head, a tapering body, and a narrow, pointed tail. In adults it's about 6 inches (15 centimeters) long but less than 2 inches (5 centimeters) wide\cite{ACS2024}.
As shown in Fig \ref{pancreas_fig}, the pancreas is an organ that sits behind the stomach, comprising both endocrine cells and exocrine cells. The majority of the pancreas is composed of exocrine cells, which form the exocrine glands and ducts. The exocrine glands produce pancreatic enzymes that are released into the intestines to help digesting foods (especially fats). Conversely, neuroendocrine cells, a distinct cell type found throughout the body, including the pancreas, secrete hormones into the bloodstream\cite{ACS2024}.
Pancreatic neuroendocrine tumors (pNETs) constitute a rare form of cancer which originate from neuro-endocrine cells. These tumors account for less than 5\% of pancreatic cancers\cite{xu2021epidemiologic} and are characterized by their slow growth, relatively more manageable treatment, and longer survival periods compared to carcinomas originating from exocrine tissue\cite{dasari2017trends}, such as pancreatic ductal adenocarcinoma (PDAC), the most common and aggressive type of pancreatic cancer. 
However, the extended incubation period of pNETs poses a significant risk to patients in advanced stages. 

Typical pNETs cases are solitary in lesion location
, though a few may be multiple\cite{low2011multimodality}. 
%These tumors can be classified into functional and non-functional types based on hormone production\cite{falconi2016enets}. Functional pNETs are often diagnosed early due to the symptoms they induce, which are typically associated with the specific hormones secreted by the tumor. In contrast, non-functional pNETs usually do not present with symptoms until the tumor becomes larger and begins to cause compression on surrounding organs, leading to more pronounced clinical symptoms\cite{halfdanarson2008pancreatic}.
\IEEEpubidadjcol
Historically, the diagnostic rate for neuroendocrine tumors has been relatively low, and they were considered rare diseases. However, with the widespread use of imageological examinations and increased rates of surgical resection, neuroendocrine tumors are no longer as rare\cite{hallet2015exploring}, underscoring the importance and non-negligibility of the task of segmenting pancreatic neuroendocrine tumors.
%apart from common PDACs
Contrast-enhanced Computed Tomography (CECT) is currently the most commonly employed medical imaging modality for staging pancreatic cancer and evaluating resectability\cite{grossberg2020multidisciplinary}. CECT can assess the vascular support of tumours by analyzing the temporal changes in attenuation within blood vessels and tissues, as captured from a rapid series of images acquired during intravenous administration of conventional iodinated contrast material\cite{miles2012current}. This process generates samples with four dimensions, comprising three spatial dimensions and multiple contrast agent diffusion phases per voxel. Each voxel in the CECT dataset corresponds to the CT number of the scan, imaging the internal anatomy of living creatures.
Accurate tumour segmentation is crucial for medical evaluation, since it facilitates subsequent tasks such as assessing resectability
\cite{tempero2017pancreatic}
and estimating overall survival\cite{skrede2020deep}\cite{jiang2021development}.

Currently, the limited availability of pNETs cases remains a challenge when training a comprehensive pancreatic cancer diagnosing model.
%The distribution of training data significantly impacts model performance. 
The existing public pancreatic tumor datasets, including the Medical Segmentation Decathlon (MSD) \cite{antonelli2022medical} and National Institutes of Health (NIH) datasets\cite{roth2016data}, only contain a small proportion of pNETs samples, making pNETs segmentation and classification a challenging task for models trained on these datasets. 
%The long-tailed distribution of samples and 
Besides, the distinct characteristics of pNETs on CECT images present significant practical challenges for the early diagnosis of pancreatic diseases. Specifically, small pNETs may not alter the morphology or contour of pancreas on CT scans and are rarely accompanied by pancreatic or bile duct dilatation. This lack of external imaging features makes the detection of tumors challenging. Moreover, pNETs are generally hyper-vascular tumors, exhibiting pronounced enhancement during the arterial phase of contrast-enhanced scans. Conversely, a smaller subset of pNETs are hypo-vascular, displaying weak enhancement, which is reminiscent of pancreatic cancer\cite{jeon2017nonhypervascular}. The small lesion size and diversity among samples make pNETs segmentation a challenging task.

Modern Artificial-Intelligence-based (AI) algorithms have revolutionized Computer-Aided Detection and Diagnosis (CAD) systems, achieving performance comparable to that of human experts in medical image analysis\cite{mckinney2020international}. 
It paves the path toward downstream tasks, including automated cancer diagnosis and prognosis prediction\cite{wang2024pathology}.
However, previous work\cite{cao2023large} based on non-specific pancreatic cancer datasets, which represent only about 1/10 of pNETs cases compared to PDACs in the training set, have shown relatively low accuracy in the task of pNETs segmentation and classification. 
%This poor performance primarily due to the rarity and diversity of pNETs, which leads to model misclassification and missed detection in cases with blurs and low image contrast.
The rarity of pNET cases and the diversity of lesion anatomy translate into low sensitivity and misclassification on non-Contrast CT scans with blurs.

To address these limitations, we introduce a well-annotated PNETs CECT dataset, consisting of 469 3D samples.
%of a total 21,122 slices. 
It was further split and processed into a total of 21,122 class-balanced slices.
Additionally, we present multi-phase well-annotated CECT imaging, including arterial phase and venous phase, potentially enriching multi-phase registration and fusion tasks with richer visual information. The dataset is based on a retrospective observational study of pNETs cases. Definitive evidence was required through biopsy pathology for diagnosis when searching the cases. The classification of the surgical pathology was determined based on the 2019 WHO classification of tumors of the digestive system\cite{nagtegaal20202019}.  A medical student manually performed slice-by-slice segmentation of the pancreas as ground-truth, which was verified/modified by an experienced radiologist.

Our main contributions can be listed as follows:
\begin{itemize}
% \item The introduction of a new PNETs CECT dataset, addressing the scarcity of research on this relatively rare tumor. 
% \item A comprehensive pipeline for PNETs segmentation through a three-stage network.
% \item 
\item We introduce a novel CECT dataset, exclusively dedicated to Pancreatic Neuroendocrine Tumors, which is characterized by its scarcity. The dataset includes multi-phase CT scans, coupled with meticulously segmented lesions at slice-by-slice level, offering a unique opportunity to address the paucity of data for this relatively rare tumor type.
\item We highlight the challenges posed by our dataset and explore the characteristics of PNETs cases within it.
\item We evaluate the performance of a suite of highly regarded U-Net-based medical image segmentation methods, including a straightforward baseline proposed by ourselves, as a benchmark for further studies.
\end{itemize}
%Our dataset will be released in the near future.

% \section{Related Work}

% \noindent\textbf{Pancreas and Pancreatic Tumor Segmentation.} 
% This segmentation task is quite challenging and interpreter-dependent because of irregular contour, small-scaled lesions, and ill-defined margins. The automated segmentation of pancreas and pancreatic tumor techniques can help radiologists make decisions more efficiently and accurately during the diagnosis and detection stages. Among deep learning approaches, CNN-based techniques stand out by their excellent feature extraction capabilities and effective learning properties. Several CNN-based methods have shown promising performance in medical image segmentation tasks and even helped models achieve state-of-the-art (SOTA) results. Diving deeper into the structures of models, U-Net architectures play a vital role, and most state-of-the-art models utilize variants of U-Net architectures. Attention U-Net \cite{oktay2018attention}, AX-Unet \cite{yang2022ax}, and nnUnet-v2 \cite{isensee2024nnunetrevisitedrigorousvalidation} have all achieved state-of-the-art performance in pancreas and pancreatic tumor segmentation tasks. The performance of these U-Net variants is further enhanced by leveraging techniques such as attention mechanisms \cite{al2023improved}, pruning schemes \cite{zhou2018unet++}, and residual architecture \cite{jha2019irnet}. However, those CNN-based approaches may struggle with images that require capturing long-range dependencies, potentially limiting their performance.

\begin{figure*}[t]
\centering
\includegraphics[width=1.0\textwidth]{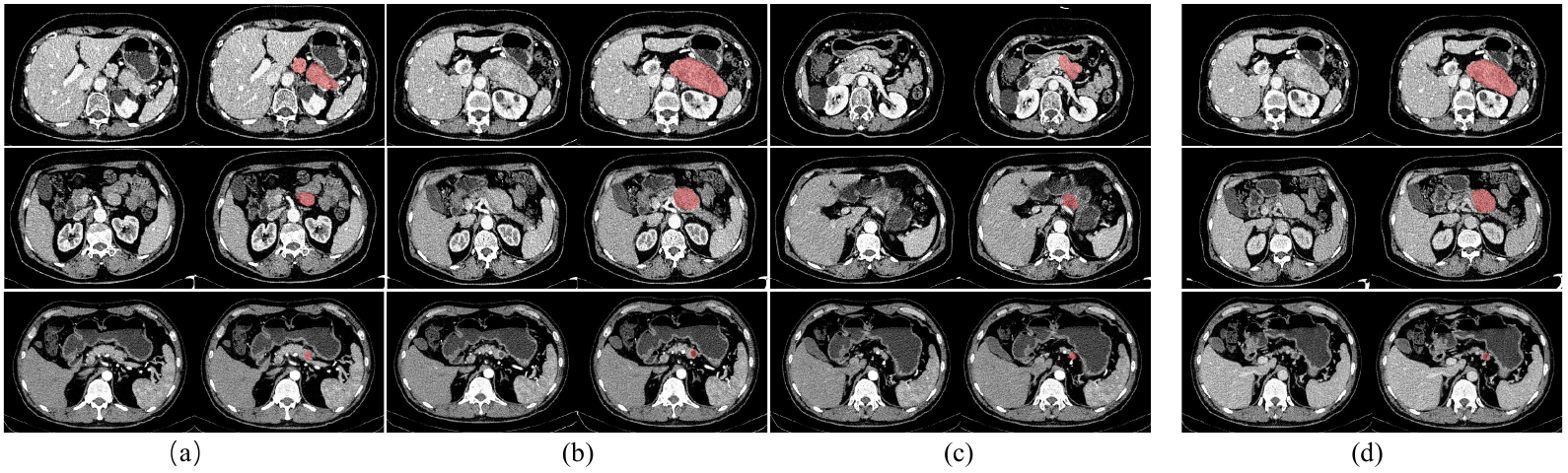} % Reduce the figure size so that it is slightly narrower than the column. Don't use precise values for figure width.This setup will avoid overfull boxes.
\caption{Examples of our pNETs dataset. Each row represents a patient's CECT image. (a), (b), (c) illustrates the image and annotated tumor area in Arterial phase, (d) illustrates the Venous phase.}
\label{dataset-2_fig}
\end{figure*}

\section{Related Datasets}
% Several publicly available datasets exist, primarily focused on the diagnosis of pancreatic cancer. However, few are suitable for lesion segmentation tasks. These datasets include diseases such as pancreatic ductal adenocarcinoma (PDAC), pancreatic neuroendocrine tumors (pNETs), and intraductal papillary mucinous neoplasm (IPMN) of the pancreas. Below, we provide a brief overview of the sources and characteristics of these datasets.

\noindent\textbf{MSD dataset\cite{antonelli2022medical}.} The MSD pancreas tumor dataset contains 420 contrast-enhanced CT scans of patients with pancreatic lesions (PDAC, PNET, and IPMN) from the Memorial Sloan Kettering Cancer Center (New York, NY, USA). Pancreas and tumor segmentation are provided for the training subset, but the exact pathology of individual tumors remains unannotated, limiting detailed clinical or diagnostic evaluations of segmentation results.

\noindent\textbf{NIH-PCT dataset\cite{roth2016data}.} NIH-PCT dataset consists of 80 portal venous phase abdominal CT scans from patients treated at the NIH Clinical Center. This dataset was curated specifically to highlight normal pancreatic morphology. Pancreas segmentations are provided in NIfTI format, annotated on a slice-by-slice basis by a medical student and subsequently refined by a radiologist. Since all CT scans in this dataset exhibit normal pancreatic morphology, they can only serve as negative samples in the context of lesion segmentation tasks.

\noindent\textbf{TCIA-PDA dataset\cite{nationalradiology}.} The TCIA-PDA dataset, developed by the National Cancer Institute’s Clinical Proteomic Tumor Analysis Consortium (CPTAC), includes 60 abdominal CT scans and 6 MRIs from patients diagnosed with pancreatic ductal adenocarcinoma (PDAC). The dataset is notable for its extensive metadata, including patient demographics, tumor characteristics, and survival outcomes, collected from consortium institutions across the USA, Canada, and Europe. However, the absence of tumor or pancreas segmentation limits the range of available downstream tasks.

% \textbf{PANORAMA study\cite{Alves2024PANORAMA}.} to do

\section{Our PNET Dataset}
We introduce the well-curated dataset for pNET and provide its basic statistics and our baseline 2-D segmentation methods.

% \subsection{Data Collection and Preparation}
\noindent\textbf{Original Volumetric Annotated Data.} 
Our pNETs dataset consists of 469 original 3-D volume data of shape $512\times~512\times~h$, each of which is a 3-D CECT result of a patient. 
$h\in[150,270]$ is the amount of the scanned transverse planes, typically covering the region from the lower chest to the bottom of the kidneys. 
The slice interval is evenly set within 1mm to 1.5mm. 
Each volumetric scan is accompanied by a ground-truth 3D tumor segmentation mask, annotated and cross-validated by expert radiologists from the Fudan University Shanghai Cancer Center. 
Figure~\ref{dataset-2_fig} showcases three examples in our dataset where we randomly select slices in Arterial and Venous phases.

\noindent\textbf{Post-processed 2-D Pancreas Images.}
We also prepare a post-process version of the dataset by segmenting and augmenting the pancreas area within each CT slice.
In detail, we apply TotalSegmentator~\cite{Wasserthal_2023}, a famous open-source 3D pancreas segmentation deep network trained on non-contrast CT scans, to obtain 3D pancreas segmentations, whose output is of the same size as the input.
Next, for each slice, we crop out a $256\times~256$ squared region whose center is the center of the bounding box of the pancreas mask on the slice.
Note that no interpolation is used in order to avoid unnecessary distortions. 
Our post-processed version helps highlight information within pancreas, which might benefit pNET segmentation. 

\begin{figure}[t]
\centering
\includegraphics[width=1.0\columnwidth]{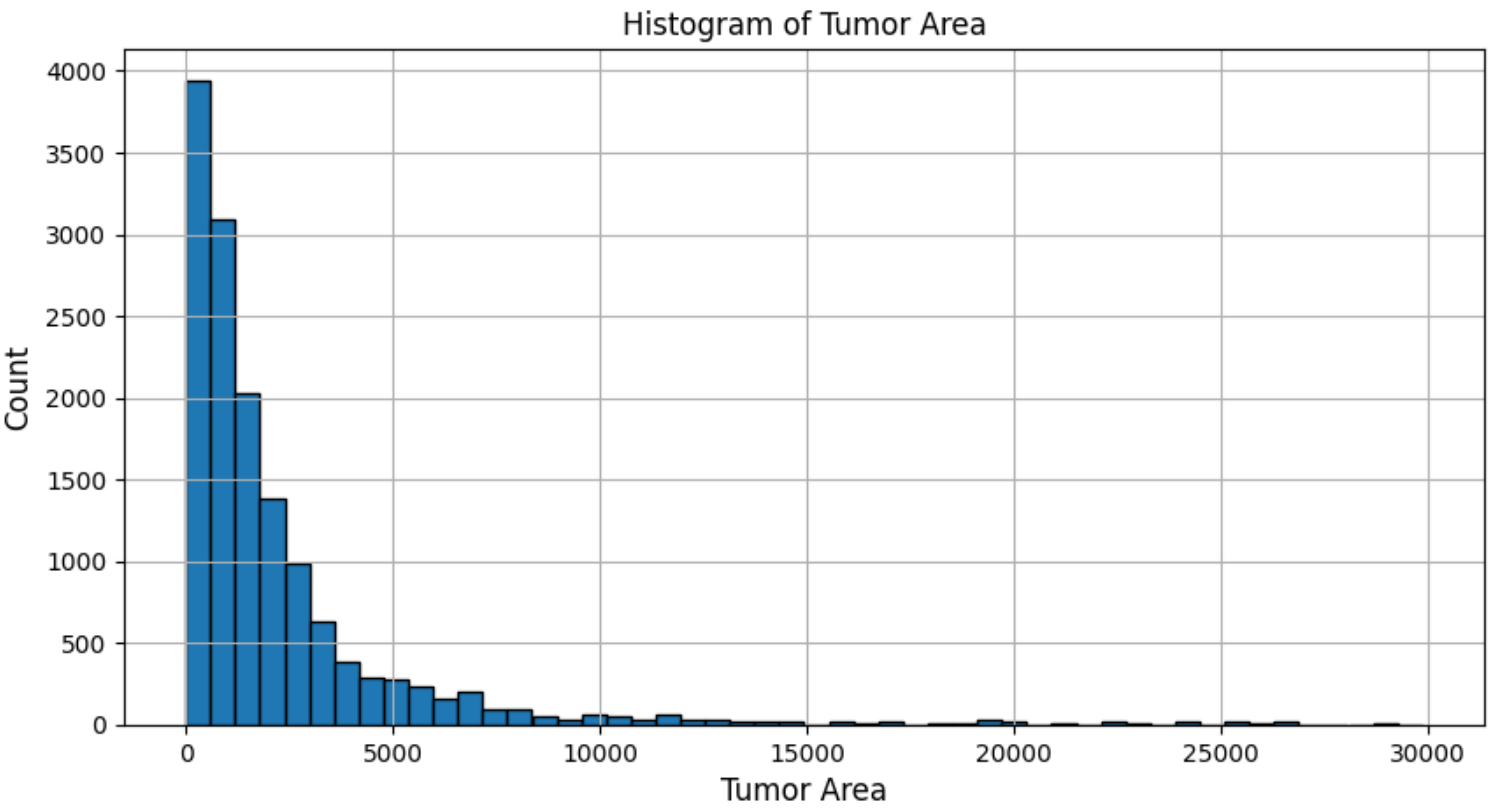} % Reduce the figure size so that it is slightly narrower than the column. Don't use precise values for figure width.This setup will avoid overfull boxes.
\caption{Histogram of tumor area for the sliced 2-D images containing tumor in our dataset. Y-axis counts the amount of pixels marked positive in the GT mask. We see that most of the images are only with a small tumor mask, resulting in a skewed histogram.}
\label{statistic}
\end{figure}

Given that tumors occupy a small portion of the pancreas, our dataset inevitably contains a large number of slices without tumors. Our statistics reveal that there are 14,501 images with tumors and 121,707 images without. To address this imbalance, we adjust the dataset to achieve a 1:1 ratio between tumor and non-tumor images, resulting in 21,122 images for training and 2,596 for validation. Additionally, we present a histogram depicting the occurrence of tumor pixels within each image, as shown in Figure~\ref{statistic}. This analysis highlights that many slices contain only a small portion of the tumor, underscoring the need for effective algorithms to handle such imbalanced data.

\begin{figure}[t]
\centering
\includegraphics[width=1.0\columnwidth,scale=2]{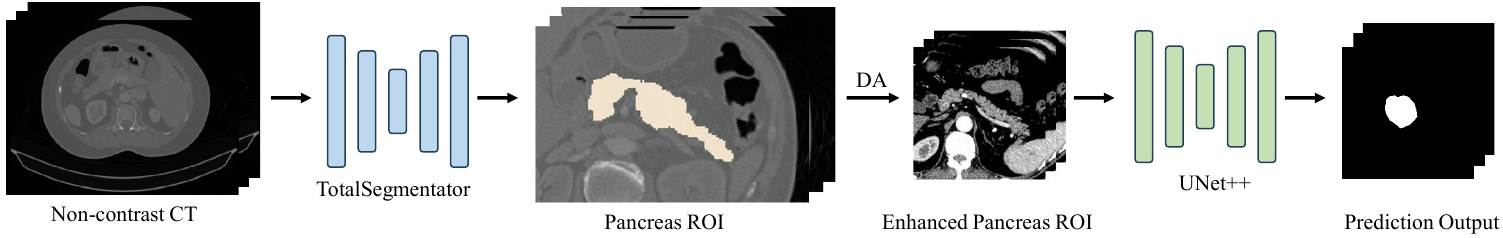} % Reduce the figure size so that it is slightly narrower than the column. Don't use precise values for figure width.This setup will avoid overfull boxes.
\caption{The framework of our proposed baseline method. For each 2D CT image, we first segment out the pancreas mask from the given CT image using pretrained models. Next, we crop and augment the area and send the corresponding image into our model for pNET segmentation.}
\label{framework_fig}
\end{figure}

\noindent\textbf{Baseline Segmentation Models.}
We depict our baseline together with the post-processing process that converts 3-D volume data to 2-D slices in Figure~\ref{framework_fig}.
First, we applied data augmentation techniques, including flipping, rotation, and random cropping for the slices containing tumors to improve the diversity of training.
Next, for 2-D tumor segmentation, we apply several classic UNet-based models as baselines, including UNet~\cite{ronneberger2015unetconvolutionalnetworksbiomedical}, UNet++~\cite{zhou2018unet++} and Attention UNet~\cite{oktay2018attentionunetlearninglook}.

\begin{figure}[t]
\centering
\includegraphics[width=1.0\columnwidth]{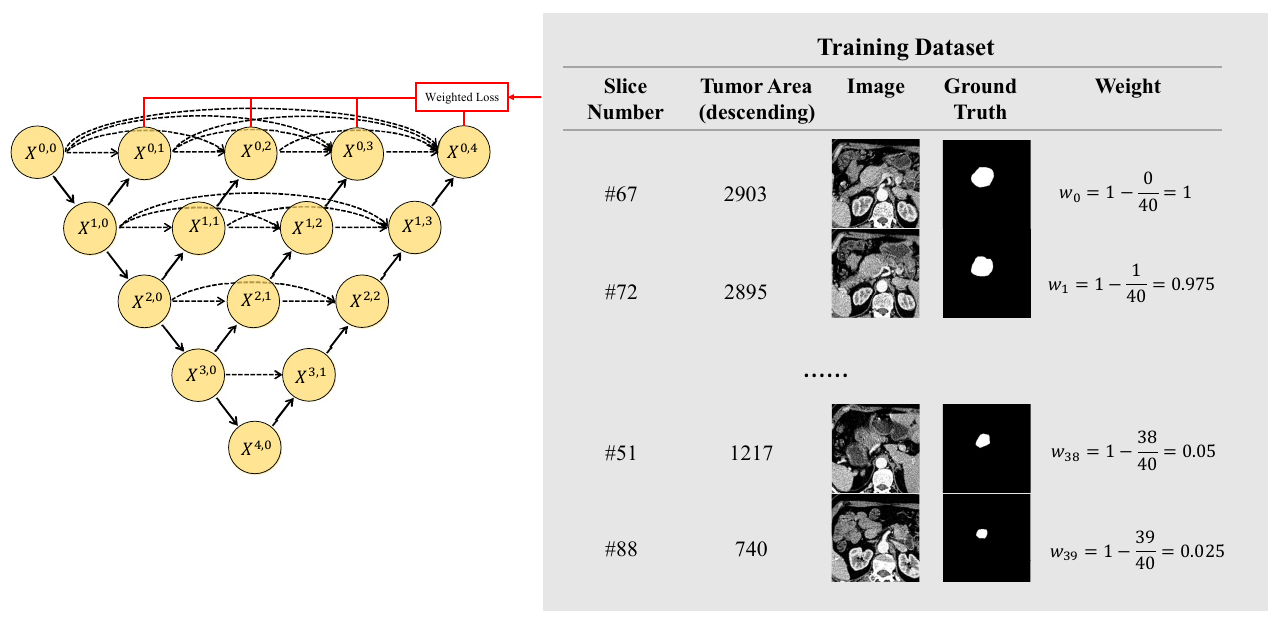} % Reduce the figure size so that it is slightly narrower than the column. Don't use precise values for figure width.This setup will avoid overfull boxes.
\caption{Our proposed slice-wise weight loss, which assigns sampling weights to each 2D CT image based on the relative ranking of its tumor area among all slices from the same patient.}
\label{loss_function_figure}
\end{figure}

% \noindent\textbf{Slice-wise Weight Loss}.
To train the networks, we propose the slice-wise weight loss which assign different loss weights to different slices, to ensure images with larger labeled tumor areas receive greater loss weights, as these samples usually contain more distinguishable tumor features. 
The strategy is depicted in Figure~\ref{loss_function_figure}. 
Concretely, each patient's volumetric data, we first rank the corresponding 2D images in descending order based on the tumor area.
Then, for each slice, we weigh the popular dice loss with the rank ratio, defined as the rank value divided by the total number of slices containing tumor. $\mathcal{L} = (1-r)\cdot \mathcal{L}_\emph{Dice}$,
where, $r$ is the rank ratio. 
The dice loss\cite{dice1945measures} $\mathcal{L}_\emph{Dice}$ is defined as follows:
\begin{equation}
\mathcal{L}_\emph{Dice} = 1 - \frac{2 \sum_{i=1}^{N} p_i g_i}{\sum_{i=1}^{N} p_i^2 + \sum_{i=1}^{N} g_i^2} 
\end{equation}
where, $p_{i}$ represents the predicted values, $g_{i}$ represents the ground truth values, and $N$ is the number of pixels.

% \subsection{Our Method}

\section{Experiment}

\noindent\textbf{Details and Evaluation Metrics.}
We trained every baseline model on one NVIDIA 4090 GPU using the Adam optimizer~\cite{kingma2014adam} with a momentum of 0.95, a learning rate of 2e-5, and a batch size of 4. The maximum training epoch is set to 100. The training of a single baseline usually lasts six hours.
We arbitrarily split all 3-D CT samples from the whole dataset into training and testing data at a ratio of $9:1$. 
To measure segmentation performance of the models, we utilize the popular Dice similarity coefficient (DCE).
% \begin{equation}
% \text{DSC} = \frac{2 |A \cap B|}{|A| + |B|}
% \end{equation}
% where \( A \) and \( B \) represent the ground truth and prediction mask separately.
% \subsection{Results and Discussions}
We compare several baselines on our dataset, including Attention UNet, UNet, and UNet++.
We also separately train these models with the proposed slice-wise weight loss to verify its effectiveness.

The experimental results are reported in two aspects, i.e., the dice of pNET on slices with pancreas (generally only a portion of these slices contain tumor), and the dice of pNET on slices with ground-truth pNET (also defined as the ROI slices).
While the latter helps analyze the overall accuracy of the detection methods, the former also help analyze the false positive on slices without pNET.
% Considering that in real-world applications, users generally have basic medical knowledge of the approximate positioning of pancreas, we do not report the pNET detection results on the slices without pancreas.

\begin{figure}[t]
\centering
\includegraphics[width=1.0\columnwidth]{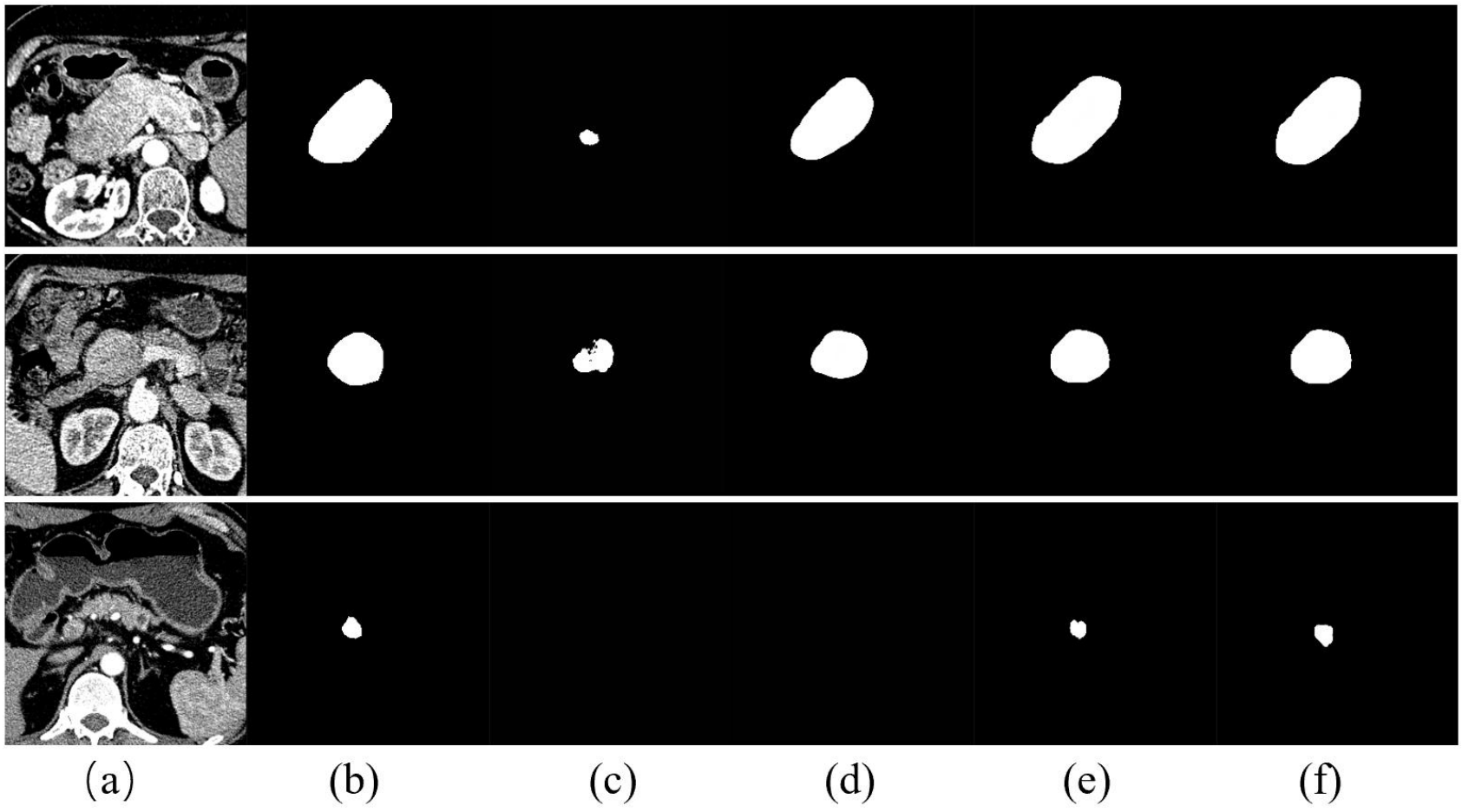} % Reduce the figure size so that it is slightly narrower than the column. Don't use precise values for figure width.This setup will avoid overfull boxes.
\caption{Segmentation examples of our method and other baselines: (a) original image, (b) ground truth, (c) Attention-UNet, (d) U-Net, (e) U-Net++, and (f) U-Net++ with slice-wise weight loss function. }
\label{result_figure}
\end{figure}

\noindent\textbf{Results and Analysis.}
In Figure \ref{result_figure}, we visualize some of the segmentation results generated by baseline methods, and in Table \ref{result_table} we report the quantitative results.
As shown in the overall results, the baseline models give decent performances on the pancreas slices, where the DCE values generally exceed 0.7. 
For the slices with no tumor where the GT mask is blank, if the models give false positives, the corresponding metric would be zero, otherwise one. 
As a result, the result indicates that the baseline models generally have relatively low false positives.
However, we see that the metrics for ROI slices are noticeably lower, and from the visualization results, the UNet baseline struggles with recalling positives from slices with small tumor GT.
% In general, UNet++ with the weighted loss function generates more accurate segmentation masks compared to the baseline methods.
% Also, the existing models perform well on large tumor slices, and our proposed loss function further enhances the precision of detecting these areas. 
Although those slices with larger tumor ground-truths might be more obvious or important for clinical diagnosis, these slices with smaller ones also could non-negligibly affect the accuracy of the decision-making and help root-causing the issues.
% which, while less clinically significant, still hold value.
Besides, the baselines using the typical segmentation architectures can be improved with the proposed slice-wise weight loss on DSC.

We leave room for more advanced segmentation methods, 1) via non-CNN architectures such as Transformer-based~\cite{vaswani2017attention} or Mamba-based~\cite{wang2024mamba} models, 2) via prompting existing large models~\cite{ma2024segment} with more prior knowledge, or 3) via 3-D CNNs, which additionally considers cross-slice information (nevertheless, the computational complexity might be much higher).
% Future work could focus on improving the detection of small slices using larger, more advanced models, potentially incorporating recent 3D medical segmentation networks. 

\begin{table}[t]
\caption{Quantitative evaluation of the baselines with/without slice-wise weight loss function}
\renewcommand{\arraystretch}{1.3}
\begin{center}
\begin{tabular}{c|c|cc}
\hline
\textbf{Slices} & \textbf{Methods} & \textbf{DSC} & \textbf{DSC(weight loss)} \\
\hline
\multirow{3}{*}{\rotatebox[origin=c]{90}{\textit{Pancreas}}} & \text{Attention UNet}& 0.694 & 0.701\\
& \text{U-Net} &  0.7521 & 0.762\\
& \text{UNet++} & 0.7361  & 0.768\\
\hline
\multirow{3}{*}{\rotatebox[origin=c]{90}{\textit{ROI}}} & \text{Attention UNet}&0.185 & 0.198
\\
& \text{U-Net} & 0.392 & 0.347 \\
& \text{UNet++} & 0.366 & 0.433\\
\hline

% \textbf{Table}&\multicolumn{3}{|c|}{\textbf{Table Column Head}} \\
% \cline{2-4} 
% \textbf{Head} & \textbf{\textit{Table column subhead}}& \textbf{\textit{Subhead}}& \textbf{\textit{Subhead}} \\
% \hline
% copy& More table copy$^{\mathrm{a}}$& &  \\
% \hline
% \multicolumn{4}{l}{$^{\mathrm{a}}$Sample of a Table footnote.}
\end{tabular}
\label{result_table}
\end{center}
\end{table}

% \subsection{Ablation Studies}
% We first evaluate the effectiveness of our designed weighted loss function. Table n shows that replacing the weighted loss with the standard Dice loss leads to a increase in DSC on our dataset. This suggests that prioritizing larger tumor areas, which typically contain more prominent features, enhances the model's ability to learn tumor characteristics. 

% We also evaluate the impact of the data augmentation strategy on model performance. As shown in Table n, incorporating data augmentation improves the DSC for images with tumors. This suggests that our data augmentation method enhances dataset diversity, particularly in tumor regions, leading to improved tumor identification.

% \begin{table}[htbp]
% \caption{Effictiveness of weighted loss function for segmentation on our dataset}
% \begin{center}
% \begin{tabular}{ccc}
% \hline
% \textbf{Architecture} & \textbf{DSC} \\
% \text{U-Net} & 0.694 \\
% \text{U-Net} &  0.7521 \\
% \text{UNet++} & 0.7361 \\
% \hline
% \textbf{Ours} & 0.768

% \textbf{Table}&\multicolumn{3}{|c|}{\textbf{Table Column Head}} \\
% \cline{2-4} 
% \textbf{Head} & \textbf{\textit{Table column subhead}}& \textbf{\textit{Subhead}}& \textbf{\textit{Subhead}} \\
% \hline
% copy& More table copy$^{\mathrm{a}}$& &  \\
% \hline
% \multicolumn{4}{l}{$^{\mathrm{a}}$Sample of a Table footnote.}
% \end{tabular}
% \label{result_table}
% \end{center}
% \end{table}

\section{Conclusion}
In this paper, we primarily introduce a well-annotated pancreatic neuroendocrine tumor (pNETs) Contrast-Enhance CT dataset.
We focuses on the rare pancreatic diseases by providing more well-curated samples compared to existing datasets, promoting further research in pancreatic tumor detection. 
As baseline detection methods,
we propose a specially designed slice-wise weight loss function for the UNet-based models, which demonstrates effective performance in managing the overwhelming detection results.
We hope the dataset also beneficial to the diagnosis of rare tumor.

\bibliographystyle{IEEEtran}
\bibliography{IEEEabrv, IEEEexample}

\vspace{12pt}

\end{document}